\documentclass[runningheads]{llncs}
\usepackage{graphicx}
\usepackage{wrapfig}
\usepackage{tikz}
\usepackage{comment}
\usepackage{amsmath,amssymb} 
\usepackage{color}
\usepackage{tabularx}
\usepackage{hyperref}       
\usepackage{url}            
\usepackage{booktabs}       
\usepackage{amsfonts}       
\usepackage{nicefrac}       
\usepackage{xcolor}         
\usepackage{colortbl}

\hypersetup{
  colorlinks,
  citecolor=blue,
  linkcolor=red,
  urlcolor=red
}
\usepackage{amssymb}
\usepackage{pifont}
\newcommand{\cmark}{\ding{51}}%
\newcommand{\xmark}{\ding{55}}%

\usepackage[accsupp]{axessibility}  

\usepackage{kotex}

\usepackage{multirow}
\usepackage{mathtools}

\begin{document}
\pagestyle{headings}
\mainmatter
\def\ECCVSubNumber{3094}  

\title{GCISG: Guided Causal Invariant Learning for Improved Syn-to-real Generalization} 

\titlerunning{ECCV-22 submission ID \ECCVSubNumber} 
\authorrunning{ECCV-22 submission ID \ECCVSubNumber} 
\author{Anonymous ECCV submission}
\institute{Paper ID \ECCVSubNumber}

\titlerunning{Guided Causal Invariant Learning for Syn-to-real Generalization}
%
\author{Gilhyun Nam\inst{1} \and
Gyeongjae Choi\inst{1} \and
Kyungmin Lee\inst{2}}
\authorrunning{G.Nam, G.Choi, and K.Lee}
%
\institute{Agency for Defense Development (ADD), Daejeon, Korea  \\
\email{\{ngh707, def6488\}@gmail.com} \\ \and
 Korea Advanced Institute of Science and Technology (KAIST), Daejeon, Korea
\email{kyungmnlee@kaist.ac.kr}}

\maketitle

\begin{abstract}
Training a deep learning model with artificially generated data can be an alternative when training data are scarce, yet it suffers from poor generalization performance due to a large domain gap. 
In this paper, we characterize the domain gap by using a causal framework for data generation.  
We assume that the real and synthetic data have common content variables but different style variables.
Thus, a model trained on synthetic dataset might have poor generalization as the model learns the nuisance style variables.
To that end, we propose causal invariance learning which encourages the model to learn a style-invariant representation that enhances the syn-to-real generalization. 
Furthermore, we propose a simple yet effective feature distillation method that prevents catastrophic forgetting of semantic knowledge of the real domain. 
In sum, we refer to our method as {Guided Causal Invariant Syn-to-real Generalization} that effectively improves the performance of syn-to-real generalization.
We empirically verify the validity of proposed methods, and especially, our method achieves state-of-the-art on visual syn-to-real domain generalization tasks such as image classification and semantic segmentation. 



\end{abstract}

\section{Introduction}

While deep neural networks have shown their great capability on various computer vision tasks, the majority of them count on a sufficient amount of training data with qualified labels. However, obtaining data or labels is expensive or difficult. Thus, a manually generated training dataset can be an alternative to the shortage or absence of training data. Also, by using the computer graphics engine~\cite{DBLP:journals/corr/HandaPBSC15a,DBLP:journals/corr/McCormacHLD16,DBLP:journals/corr/abs-1708-05869}, one can obtain labels without any cost of human labor~\cite{gaidon2016virtual,RosCVPR16}. 

However, training with synthetic data cannot fully replace the real training data as they suffer from poor generalization performance in the real domain. 
Many previous works explain the reason for inferior performance by the existence of a large domain gap~\cite{Maximov_2020_CVPR,Tremblay_2018_CVPR_Workshops} which makes the model overfits to the synthetic domain. 
In this paper, we break down the domain gap between the synthetic and real data by using the structural causal model. 
We build a causal model that an image is generated from two latent variables: style variables such as texture and content variables such as shape. 
We assume that the synthetic and real data differ in style variables while having common content variables that are relevant to the downstream tasks. Furthermore, we also assume that the model learns the task-irrelevant style variables, therefore exhibiting poor generalization on the real domain. 
The recent studies~\cite{baker2018deep,geirhos2018imagenet} show that convolutional neural networks tend to rely on texture rather than shape which supports our claim.

To that end, we propose causal invariant learning for syn-to-real generalization by promoting causal invariance loss to learn style-invariant representation. 
Our method is related to the contrastive learning method, which learns representation by aligning positive pairs. 
We apply data augmentation to generate a pair of images with the same content but different style variables, then align their distributions of representations to achieve style-invariance. Our loss function is similar to that of~\cite{zheng2021ressl,fang2021seed,xu2020knowledge,park2019relational}, where those methods are for other than syn-to-real generalization such as self-supervised learning or knowledge distillation. 

Furthermore, many methods show that using real-domain guidance with ImageNet pre-trained model for regularization is effective for syn-to-real generalization~\cite{asg,csg}. 
Chen et al.~\cite{asg} utilized the knowledge distillation loss by re-using the pre-trained ImageNet classifier and Chen et al.~\cite{csg} used intermediate features of the ImageNet pre-trained network to compute negative samples in contrastive learning.
Those methods posit that the ImageNet classification is strongly correlated to the syn-to-real generalization performance.
Instead, we propose to directly guide the model by simply regularizing the feature distance with the ImageNet pre-trained model, yet the key is to extract the semantic knowledge by using self-attention pooling. 
We empirically show that the simple guidance loss with self-attention pooling successfully guides the model to localize the object in the image, and helps the syn-to-real generalization. 

In summary, we propose \emph{Guided Causal Invariant Syn-to-real Generalization} (GCISG), where we propose causal invariance loss that helps to learn style-invariant features and guidance loss with self-attention pooling that helps extract semantic information from ImageNet pre-trained neural network. Our contributions are listed below:
\begin{itemize}
\item By adopting the causal framework for a syn-to-real setup, we propose causal invariance loss which regularizes the model to learn a style-invariant representation that is relevant to syn-to-real generalization. 

\item We present a simple and effective guidance loss that guides the model with real-domain guidance from ImageNet pre-trained model by utilizing self-attention pooling.

\item We empirically show that our method significantly outperforms competing syn-to-real generalization methods on various vision tasks such as image classification, semantic segmentation, and object detection.

\end{itemize}

\section{Related Work}

\begin{figure}[t]
\centering
\includegraphics[width=0.95\textwidth]{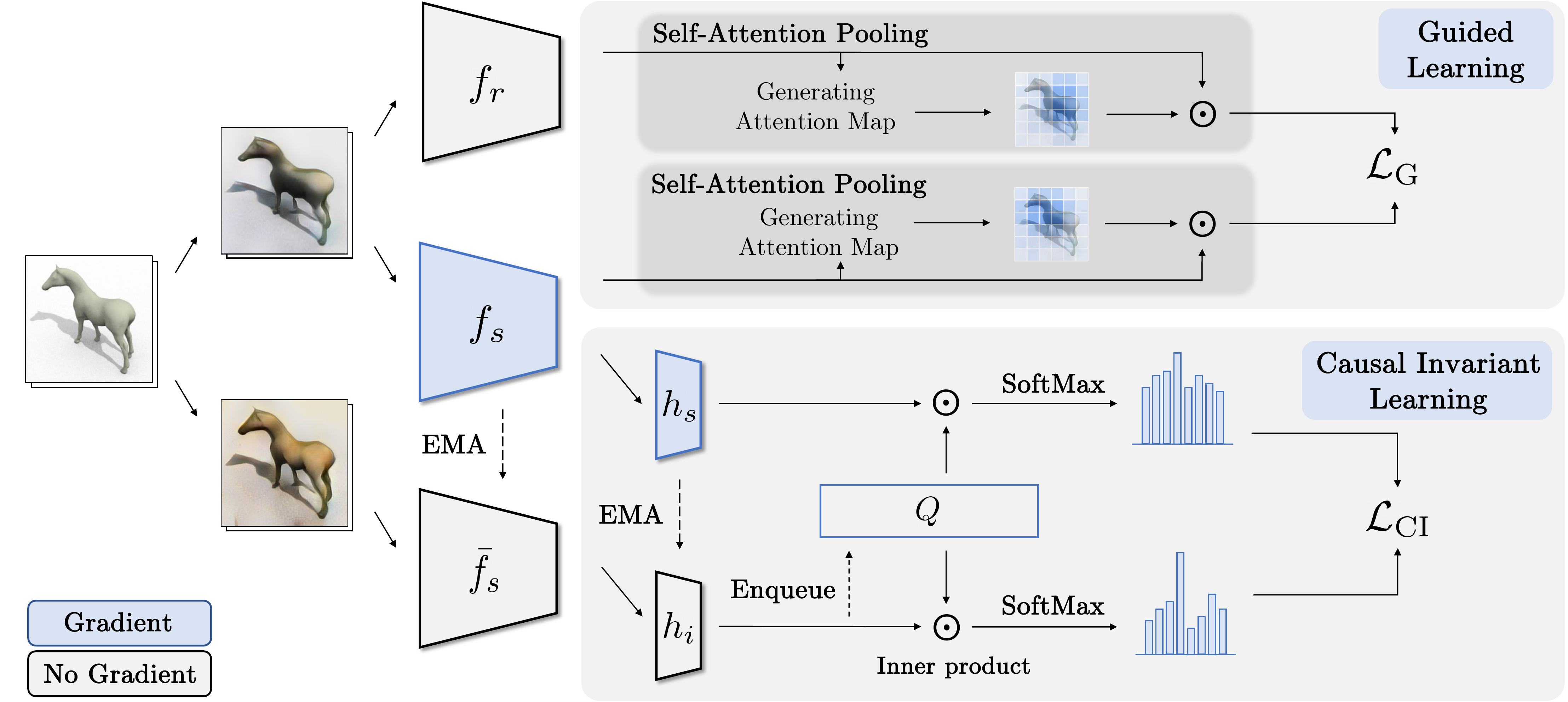}
\caption{
Overall demonstration of proposed GCISG. There are two branches: 1) the guidance loss that preserves semantic information from ImageNet pre-trained model by using self-attention pooling, and 2) the causal invariance loss that learns style-invariant representation using contrastive learning framework. 
}
\label{fig:framework}
\end{figure}

\subsubsection{Domain generalization.}
The domain generalization problem aims to train on the source domain and generalize to an unseen target domain. 
The biggest concern of domain generalization is that there is a large domain gap between source and target domains, which deteriorates the performance. 
Therefore, many methods proposed an adversarial learning framework to match the feature distributions with a prior distribution~\cite{li2018domain} or diversify the style of the source domain to enhance the generalization~\cite{yue2019domain,wang2021learning}.
Others focused on the effect of batch normalization on domain generalization. Pan et al.~\cite{Pan_2018_ECCV} showed that using instance normalization can boost the generalization, and RobustNet~\cite{Choi_2021_CVPR} used instance selective whitening to improve the robustness.  
Recent studies showed that using the knowledge from the real domain is effective for syn-to-real domain generalization problems~\cite{csg,asg,chen2018road}, and our work is concurrent to those approaches. 

{
\subsubsection{Contrastive learning and style invariance.}
Recently, contrastive learning with multi-view data augmentation has shown their efficacy in self-supervised representation learning~\cite{Wu_2018_CVPR,moco,simclr,mitrovic2020representation} and domain generalization~\cite{mahajan2021domain}.
Our contrastive learning objective is similar to that of SEED~\cite{fang2021seed}, ReSSL~\cite{zheng2021ressl}, and RELIC~\cite{mitrovic2020representation}, while SEED uses the contrastive objective to distill the representational knowledge on smaller networks, and ReSSL and RELIC are for self-supervised representation learning. 
Remark that CSG~\cite{csg} also uses contrastive learning for syn-to-real generalization, while it is different from ours as they compute contrastive loss across ImageNet pre-trained networks, while ours aim to learn style-invariant representation on the synthetic data. 
}

\subsubsection{Learning with guidance.} 

Training a model by guidance with a pre-trained model's knowledge is not only effective for syn-to-real generalization but also for various machine learning tasks such as knowledge distillation~\cite{hinton2015distilling,tung2019similarity} and incremental learning~\cite{li2017learning,castro2018end}. 
In syn-to-real generalization or adaptation, models are enforced to retain the knowledge from the ImageNet pre-trained networks by minimizing the feature distance~\cite{chen2018road}, matching the outputs of ImageNet classification~\cite{asg}, or using the contrastive learning objective~\cite{csg}. 
Similarly, the knowledge distillation methods use similar tactics to compress the knowledge from bigger networks to smaller ones~\cite{tian2019contrastive,xu2020knowledge}. 
And for incremental learning, similar methods were used to prevent catastrophic forgetting~\cite{rebuffi2017icarl}.

Our method uses self-attention pooled feature minimization to guide the model to enhance the generalization on a real domain. 
Transferring the knowledge by using attention has been also studied for knowledge distillation~\cite{zagoruyko2016paying}, and incremental learning~\cite{dhar2019learning}. While Chen et al.~\cite{csg} used attention pooling for syn-to-real generalization, there are differences in that they pool features for contrastive learning, while our objective directly minimizes the distance between them.

\section{Proposed Method}

\subsection{Causal Invariance for Syn-to-real Generalization}\label{sec:3.1}

\begin{wrapfigure}{R}{.4\columnwidth}
    \centering
    \includegraphics[width=\linewidth]{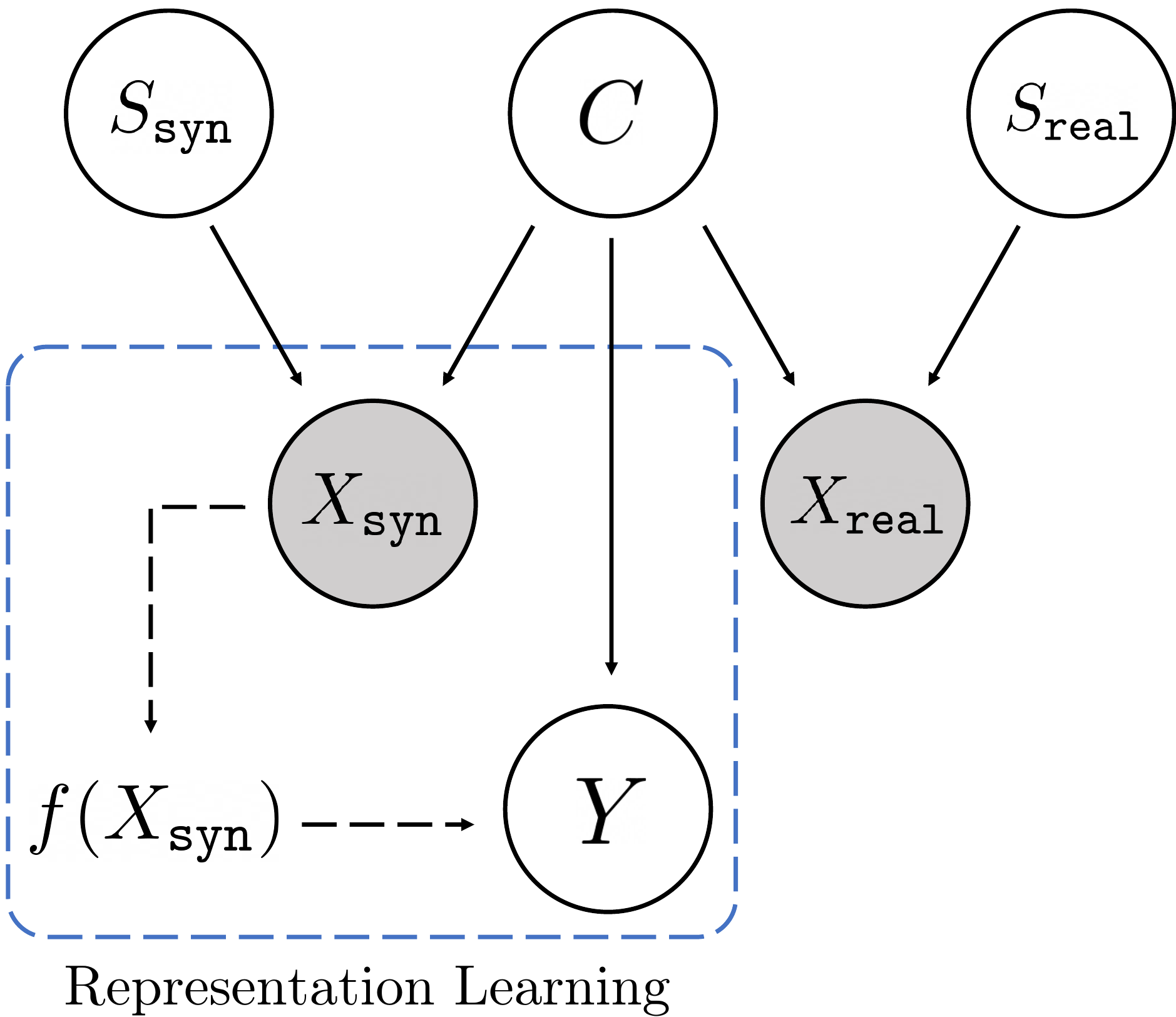}
    \caption{Causal graph for synthetic and real data. }
    \label{fig:scm}
\end{wrapfigure}
We consider a causal model for synthetic and real data generation. 

In Fig.~\ref{fig:scm}, we assume that the synthetic data $X_{\tt{syn}}$ is generated from the content variable $C$ and style variable $S_{\tt{syn}}$, and a real data $X_{\tt{real}}$ is generated from the content variable $C$ and style variable $S_{\tt{real}}$. 
Thus, we claim that the domain gap occurs because of the difference between the latent style variables of the two domains. 
We also assume that the task label $Y$ is only dependent on the content variable.
Those assumptions give us insight into why the model shows inferior generalization when solely trained with synthetic data: the model learns nuisance style variables irrelevant to the task. 
Also, a previous study~\cite{geirhos2018imagenet} corroborates our hypothesis that the convolutional neural networks are biased toward texture (i.e. the style variable). 

If one can extract style-invariant or content-preserving representation from the synthetic data, the model can enhance its generalization capability in the real domain. 
Thus, our goal is to extract style-invariant representation which is useful in the generalization on the real domain. 
To that end, we propose causal invariant syn-to-real generalization which is composed of two folds: first, we further diversify the style variable by using a strong data augmentation module such as RandAugment~\cite{cubuk2020randaugment}, and second, we use a contrastive learning framework to learn style-invariant representations.

For the synthetic training model $f_s$, we attach projector $h_s$ to project onto a smaller dimension as illustrated in Fig.~\ref{fig:framework}. For projector, we use two-layer multi-layer perceptron~(MLP) as used in many contrastive representation learning frameworks~\cite{simclr,mocov2}. 
Let us denote the encoder by $g_s = h_s \circ f_s$ and we build target encoder $\bar{g}_s = \bar{h}_i\circ \bar{f}_s$.
The weights of $\bar{g}_s$ are updated by the exponential moving average of the weight of the encoder $g_s$. 
Given a synthetic data $x$, let $x_1$ and $x_2$ be the two augmented views of $x$. 
Denote $z_1 = g_s(x_1) / \|g_s(x_1)\|_2\in\mathbb{R}^d$ 
and $\bar{z}_2 = \bar{g}_s(x_2) / \|\bar{g}_s(x_2)\|_2\in\mathbb{R}^d$ be the outputs of each encoder and target encoder with $\ell_2$ normalization. 
To achieve stable causal invariance, we aim to match the distributions of $z_1$ and $\bar{z}_2$ with enough amount of support samples, which is a similar approach to prior works in other tasks~\cite{fang2021seed,mitrovic2020representation,zheng2021ressl}. We pertain support samples $Q\in\mathbb{R}^{K\times d}$, and aim to minimize the probabilistic discrepancy between the relational distributions $p(z_1;Q)$ and $p(\bar{z}_2;Q)$. 
\begin{align}
    p_\tau(z;Q)[k] \coloneqq \frac{\exp\left(z^\top q_k/\tau\right)}{\sum_{k^\prime=1}^K \exp\left(z^\top q_{k^\prime}/\tau\right)},
\end{align}
where $q_k$ is the $k$-th component of support samples $Q$ and $\tau>0$ is a temperature hyperparameter. 
Then we define the causal invariance loss between $z_1$ and $\bar{z}_2$ by the cross-entropy loss between the relation similarities as following:
\begin{align}
    \ell_{\tt{CI}}(z_1, \bar{z}_2) = -\sum_{k=1}^K p_{\bar{\tau}}(\bar{z}_2;Q)[k] \log \big(\,p_{\tau}(z_1;Q)[k]\big), 
\end{align}
where we use distinct $\tau, \bar{\tau}>0$ to regulate the sharpness of distributions. 
For computational efficiency, we symmetrically compute compute between $\bar{z}_1 = \bar{g}_s(x_1) / \|\bar{g}_s(x_1)\|_2 $ and $z_2 = g_s(x_2) / \|g_s(x_2)\|_2$, and the total causal invariance loss for a data $x$ is given by
\begin{align}
    \mathcal{L}_{\tt{CI}}(x) = \frac{1}{2}\ell_{\tt{CI}}(z_1, \bar{z}_2) + \frac{1}{2}\ell_{\tt{CI}}(z_2, \bar{z}_1).
\end{align}
The support samples $Q$ are managed by a queue, where we enqueue the outputs of the target encoder at each iteration, and dequeue the oldest ones.
We set $K$ to $65536$ to ensure the relational distribution contains sufficient information to learn style-invariant representations.
Remark that the temperature parameter greatly affects the stability of the learning with causal invariance loss.
Generally, we choose $\bar{\tau}$ to be smaller than $\tau$, so that the target distribution is sharper. 

\subsubsection{Dense causal invariant learning for semantic segmentation.}
For the semantic segmentation task, an image contains various objects and semantic features. Therefore, we use dense causal invariance loss where we compute the loss over the patches of the representations. 
Following the method in \cite{csg}, we crop the feature maps into $N_l$ patches and pass them forward to the projector to compute causal invariance loss on each cropped representation:
\begin{align}
    \mathcal{L}_{\tt{CI-Dense}}(x) =  \frac{1}{N_l}\sum_{n=1}^{N_l} \frac{1}{2}\ell_{\tt{CI}}\left(z_{1,n}, \bar{z}_{2,n}\right) + \frac{1}{2}\ell_{\tt{CI}}\left(z_{2,n}, \bar{z}_{1,n}\right),
\end{align}
where each $z_{i,n}, \bar{z}_{i,n}$ is the encoder output of $n$-th feature map.
In our experiments, we use $N_l=8\times 8$. 

\subsection{Guided Learning for Syn-to-real Generalization}\label{sec:3.2}

In this section, we present a simple, yet effective method to guide a model with real-domain guidance from the ImageNet-pretrained model by using feature distillation.
In contrast to previous approaches such as ASG~\cite{asg} and CSG~\cite{csg}, our method does not require any task-specific information or complicated loss function. 

\subsubsection{Pooled feature distillation.}
\begin{figure}[t]
\centering
\includegraphics[width=0.95\linewidth]{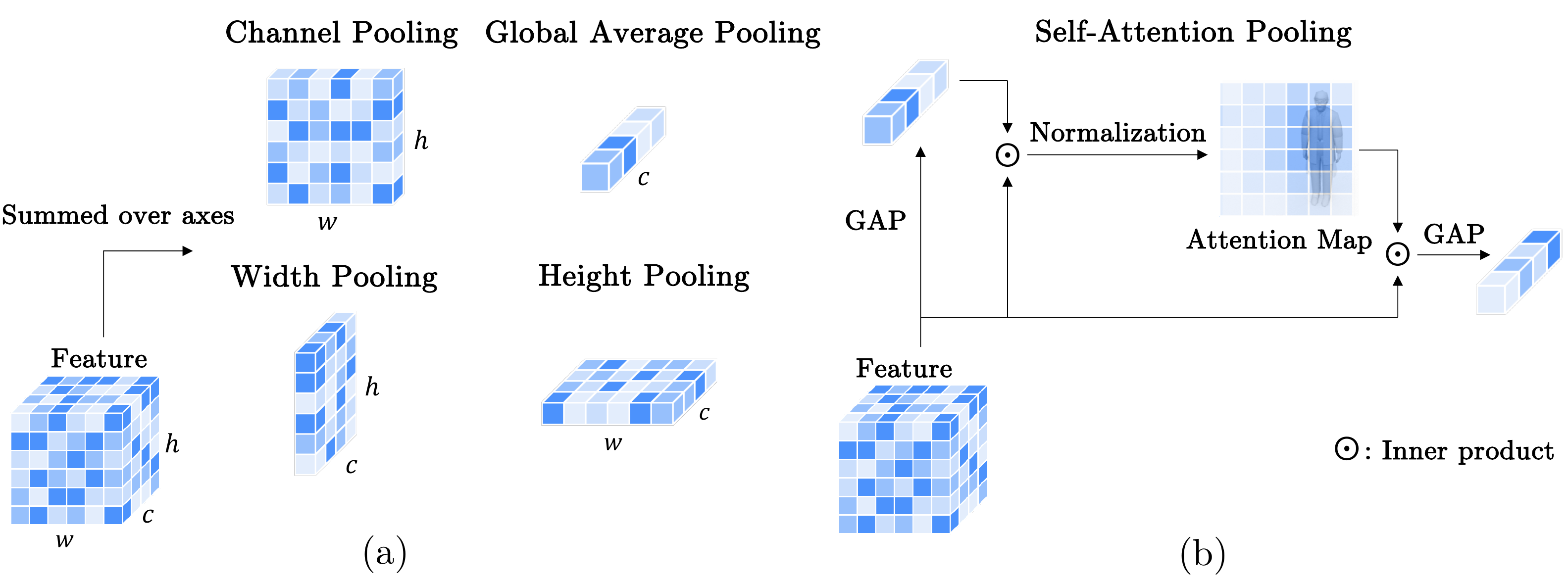}
\caption{(a) Example images of channel, global average, width and height pooled feature map. (b) Mechanism of self-attention pooling for guidance loss.}
\label{fig:vis_pooling}
\end{figure}
For a data $x$, let $f_s(x), f_r(x)\in\mathbb{R}^{C\times H\times W}$ be the feature map of an intermediate convolutional layer of ImageNet pre-trained network $f_r$ and training network $f_s$. 
Then it is straightforward to minimize the squared distance between the feature maps $f_s(x)$ and $f_r(x)$ for guidance loss to retain the knowledge of $f_r$ as much as possible while learning task-related knowledge from the synthetic data. 
To design the guidance loss, there are two opposite concepts that must be considered: rigidity and plasticity. 
The rigidity is on how much the model can retain the knowledge of $f_r$, and the plasticity accounts for the flexibility of the model to learn from the synthetic domain. 

While the direct minimization with the feature maps has high rigidity, it has low plasticity that the model cannot sufficiently learn task-related knowledge with synthetic data. 
Douillard et al.~\cite{douillard2020podnet} showed that by using an appropriate pooling operator, the model can balance the rigidity and plasticity to conduct incremental learning.
Similarly, we use the pooling operator to extract sufficient information from the feature maps and balance the rigidity and plasticity to improve syn-to-real generalization. 

In Fig.~\ref{fig:vis_pooling}~(a), we list various pooling methods that we consider in this paper. Let $P$ be such pooling operators in Fig.~\ref{fig:vis_pooling}~(a), then the guidance loss is given by the $\ell_2$ distance between the normalized pooled feature maps of $f_s(x)$ and $f_r(x)$:
\begin{align}
    \mathcal{L}_{\tt{G}}(x;P) = \left\| \frac{P(f_s(x))}{\|P(f_s(x))\|_2} - \frac{P(f_r(x))}{\|P(f_r(x)\|_2}\right\|_2^2
\end{align}

\subsubsection{Improved guidance by self-attention pooling.}
Alternatively, we propose self-attention pooling based guidance loss which improves syn-to-real generalization. 
The self-attention pooling captures semantically important features by multiplying the importance weight on each feature map. 
Therefore, the guidance with self-attention pooling allows the model to learn the semantically important feature that $f_r$ focuses on. 
The self-attention pooling is demonstrated in Fig.~\ref{fig:vis_pooling}~(b). 
Let $v\in\mathbb{R}^{C\times W\times H}$ be a feature map, then the attention map $a \in \mathbb{R}^{W\times H}$ is computed by the normalization on the global average pooling of $v$: 
\begin{align}
    a[w,h] = \frac{\sum_{c=1}^C v[c,w,h] g[c]}{\sum_{w'=1}^W\sum_{h'=1}^H \sum_{c=1}^C v[c,w',h']g[c]},
\end{align}
where $g[c] = \frac{1}{HW}\sum_{w=1}^W\sum_{h=1}^H v[c,w,h]$ is global average pooling of $v$. 
Then the self-attention pooling operator $P_a$ outputs an importance-weighted feature map with weights given by the attention map:
\begin{align}
    P_{a}(v)[c] = \sum_{w=1}^W\sum_{h=1}^H v[c,w,h]a[w,h].
\end{align}
Then our final guidance loss with self-attention pooling operator $P_a$ is given by
\begin{align}
    \mathcal{L}_{\tt{G}}(x;P_a) = \left\| \frac{P_a(f_s(x))}{\|P_a(f_s(x))\|_2} - \frac{P_a(f_r(x))}{\|P_a(f_r(x)\|_2}\right\|_{2}^{2}
\end{align}
The empirical analysis on the effect of pooling on the guidance loss is explored in section~\ref{sec:ablation}.

\subsection{Guided Causal Invariant Syn-to-real Generalization}
In this section, we present the overall description of our method for syn-to-real generalization, which we refer to \emph{Guided Causal Invariant Syn-to-real generalization}~(GCISG). 
Given a task loss $\mathcal{L}_{\tt{task}}$ that takes synthetic data $x$ with corresponding label $y$, the GCISG adds guidance loss $\mathcal{L}_{G}$ that regularizes the model to retain the semantic knowledge of ImageNet pre-trained model and causal invariance loss $\mathcal{L}_{\tt{CI}}$ to achieve style-invariant representation that promotes generalization on a real domain. Thus, the overall loss is computed by following:
\begin{align}
    \mathcal{L}(x) = \mathcal{L}_{\tt{Task}}(x,y) + \lambda_{\tt{G}}\mathcal{L}_{\tt{G}}(x) + \lambda_{\tt{CI}}\mathcal{L}_{\tt{CI}}(x).
\end{align}
Remark that our GCISG framework is agnostic to tasks, as guidance loss and causal invariance loss are computed in an unsupervised manner. 

\subsubsection{Stage-wise loss computation.}
The causal invariance loss $\mathcal{L}_{\tt{CI}}$ and the guidance loss $\mathcal{L}_{\tt{G}}$ can be computed for each intermediate layer. Denote $f^{(l)}$ to be $l$-th layer of the neural network, (e.g. the block layers of ResNet~\cite{he2016deep}), then the GCISG is computed for each layer by following:
\begin{align}
    \mathcal{L}(x) = \mathcal{L}_{\tt{Task}}(x,y) + \sum_{l} \lambda_{\tt{G}}^{(l)}\mathcal{L}_{\tt{G}}^{(l)}(x) + \sum_{l}\lambda_{\tt{CI}}^{(l)} \mathcal{L}_{\tt{CI}}^{(l)}(x).
\end{align}
In section~\ref{sec:ablation}, we conduct an ablation study on the effect of choosing layer. 

\section{Experiments}

In this section, we empirically validate the effectiveness of our method on various syn-to-real generalization tasks. We report the performance of a synthetic-trained model in the unseen real domain for evaluation. 
Furthermore, we analyze the model with various auxiliary evaluation metrics to support our claims.

\subsubsection{Evaluation metrics for causal invariance.}
To quantify the style-invariance of a representation, we introduce \emph{match rate}~($\mathcal{M}$), which evaluates the consistency of a model on images that have the same semantics but different styles.
For each image, we generate another stylized image and check if it produces the same output as the original one.
Formally, we report the match rate by the number of consistent samples out of the number of the entire validation set. 
To generate stylized images, we use photometric transforms such as Gaussian blurring and color jittering as done in~\cite{Choi_2021_CVPR}.

\subsubsection{Evaluation metrics for guidance.}
We present two different measures to quantify the quality of guidance from the ImageNet pre-trained model. 
First, we bring the linear classification layer of the pre-trained ImageNet model and attach it to the synthetic-trained model.  
Then we report the ImageNet validation accuracy ($\text{Acc}_{\tt{IN}}$) by inferring on the ImageNet validation dataset. 
Second, we evaluate the similarity between ImageNet pre-trained model and the synthetic-trained model by using \emph{centered kernel alignment}~(CKA) similarity~\cite{kornblith2019similarity}.
Here, the high ImageNet validation accuracy demonstrates that our model avoids the catastrophic forgetting of the task information, and the high CKA similarity proves the preservation of representational knowledge of a real pre-trained model. 
For the computation of CKA similarity, we collect the features of the penultimate layers of the ImageNet pre-trained model and synthetic-trained model on the validation dataset. The detailed implementations are in the supplementary material.

\subsection{Image Classification}

\begin{table}[t]
    \begin{minipage}[t]{.47\linewidth}
        \centering
        \caption{Top-1 accuracy~(\%) on VisDA-17~($\tt VisDA$) and ImageNet~($\tt IN$) validation datasets of various syn-to-real generalization methods. 
        All methods used ResNet-101 as base architecture. }
        \resizebox{0.98\textwidth}{!}{
            \begin{tabular}{@{} lcccc @{}}
            \toprule
            Method && $\text{Acc}_{\tt{VisDA}}$ && $\text{Acc}_{\tt{IN}}$\\
            \midrule
            Oracle on ImageNet  && 53.3 && \textbf{77.4}\\
            Vanilla L2 distance && 56.4 && 49.1 \\
            ROAD~\cite{zenke2017continual} && 57.1 && \textbf{77.4}\\
            SI~\cite{chen2018road} && 57.6 && 53.9\\
            ASG~\cite{asg} && 61.1 && 76.7 \\
            CSG~\cite{csg} && 64.1 && 73.8 \\
            \midrule
            \textbf{GCISG} && \textbf{67.5} && 75.4 \\
            \bottomrule
            \label{table:classification_main}
            \end{tabular}
        }
    \end{minipage}%
    \hfill
    \begin{minipage}[t]{.51\linewidth}
      \centering
      \caption{Top: Comparison with CSG and oracle on ImageNet by evaluation metrics for invariance ($\mathcal{M}$) and guidance ($\text{Acc}_{\tt{IN}}$, CKA). The oracle freezes the backbone and fine-tune the classification head. Bottom: Effects of $\mathcal{L}_{\tt{G}}$ and $\mathcal{L}_{\tt{CI}}$ in GCISG.}
        \resizebox{0.98\textwidth}{!}{
            \begin{tabular}{lcccccc}
                \toprule
                Method & $\mathcal{L}_{\tt{G}}$    & $\mathcal{L}_{\tt{CI}}$     & $\text{Acc}_{\tt{VisDA}}$  & $\mathcal{M}$  & $\text{Acc}_{\tt{IN}}$       & CKA    \\
                \midrule
                Oracle & \xmark & \xmark & 53.3    & 69.5      & 77.4           & 1.0            \\
                CSG~\cite{csg}    & \cmark & \xmark & 64.1    & 77.0      & 73.8           & 0.89         \\
                \midrule
                GCISG  & \xmark & \xmark & 55.9   & 73.3        & 52.7           & 0.72         \\
                       & \xmark & \cmark & 58.6   & 74.4        & 51.9           & 0.73           \\
                       & \cmark & \xmark & 64.3   & 76.6      & \textbf{75.9} & \textbf{0.95}        \\
                       & \cmark & \cmark & \textbf{67.5} & \textbf{78.6} & 75.4          & 0.93   \\
                \bottomrule
            \label{table:classification_metric}
            \end{tabular}
        }
    \end{minipage} 
\end{table}

\subsubsection{Datasets.}
We demonstrate the effectiveness of GCISG on VisDA-17~\cite{peng2017visda} image classification benchmark. 
The synthetic training set consists of the images rendered from 3D models with various angles and illuminations. 
The real validation set is a subset of MS-COCO~\cite{lin2014microsoft}. 
Each dataset contains the same 12 object categories.

\subsubsection{Implementation details.} 
We adopt ImageNet pre-trained ResNet-101 as the backbone. 
We use an SGD optimizer with a learning rate of 0.001, a batch size of 64 for 30 epochs. 
For data augmentations, we use the RandAugment~\cite{cubuk2020randaugment} following the protocol of CSG~\cite{csg}. 
The momentum rate is 0.996 and the number of the support sample is 65536 throughout the training.

\subsubsection{Results.}

In Table~\ref{table:classification_main}, we report the top-1 validation accuracy on VisDA-17 ($\text{Acc}_{\tt{VisDA}}$) of synthetic-trained model with various syn-to-real generalization methods. We also report the ImageNet validation accuracy ($\text{Acc}_{\tt{IN}}$) of the same synthetic-trained model as a guidance metric.
We observe that GCISG outperforms the competing methods on VisDA-17 on a large margin. 
As shown in Table~\ref{table:classification_main}, there is a correlation between the ImageNet classification accuracy and syn-to-real generalization performance. 
However, the correlation is not strong that the methods with higher ImageNet accuracy such as ASG~\cite{asg} and ROAD~\cite{chen2018road}, do not necessarily have higher validation accuracy on VisDA-17.
Thus, sufficiently high ImageNet validation accuracy might be useful for syn-to-real generalization, but it may hamper the generalization if it is too much as it transfers knowledge that is overfitted to the ImageNet classification task.

Table~\ref{table:classification_metric} shows that our method not only achieves a higher VisDA-17 validation accuracy but also has a higher match rate ($\mathcal{M}$) and CKA similarity than CSG.
The higher match rate indicates that the learned representation is invariant to the style change, and the higher CKA similarity indicates the conformation of learned representation to the ImageNet pre-trained network. 
Therefore, one can observe that both factors have positive impacts on syn-to-real generalization.

Table~\ref{table:classification_metric} reports the effect of using causal invariance loss and guidance loss on VisDA-17 validation accuracy and auxiliary evaluation metrics. 
Remark that CSG~\cite{csg} applied guidance loss in contrastive learning with ImageNet pre-trained features. We have updated state-of-the-art performance only with our guided learning without using the complex siamese network and contrastive loss as used in CSG~\cite{csg}.
We observe that the guidance loss generally boosts ImageNet validation accuracy and CKA similarity. 
This clearly demonstrates that the transferring knowledge from the ImageNet pre-trained model which holds useful knowledge of real domain helps the syn-to-real generalization. 
Moreover, using the causal invariance loss generally boosts both match rate and VisDA-17 validation accuracy. 
This also demonstrates that the style-invariance of the model enhances the syn-to-real generalization.

\subsection{Ablation Study} \label{sec:ablation}

\begin{figure}[t]
    \begin{minipage}[t]{.55\linewidth}
      \centering
        \includegraphics[width=\linewidth]{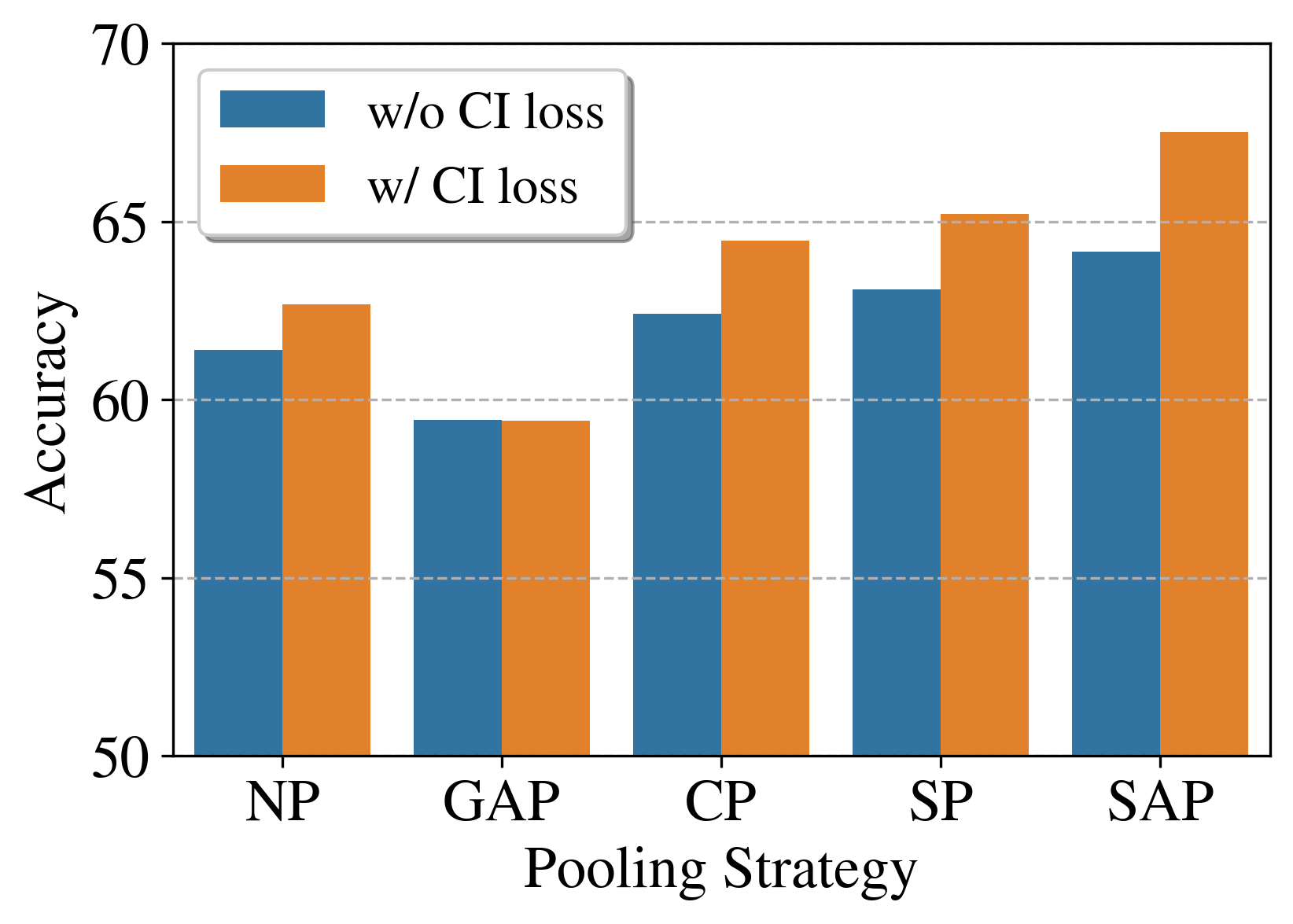}
        \caption{Ablation of feature pooling strategies for $\mathcal{L}_\text{G}$. NP, GAP, CP, SP, and SAP denote no pooling, global average pooling, channel pooling, spatial pooling, and self-attention pooling respectively.}
        \label{fig:pooling_chart}
    \end{minipage}%
    \hfill
    \begin{minipage}[t]{.40\linewidth}
      \centering
      \centering
        \includegraphics[width=\linewidth]{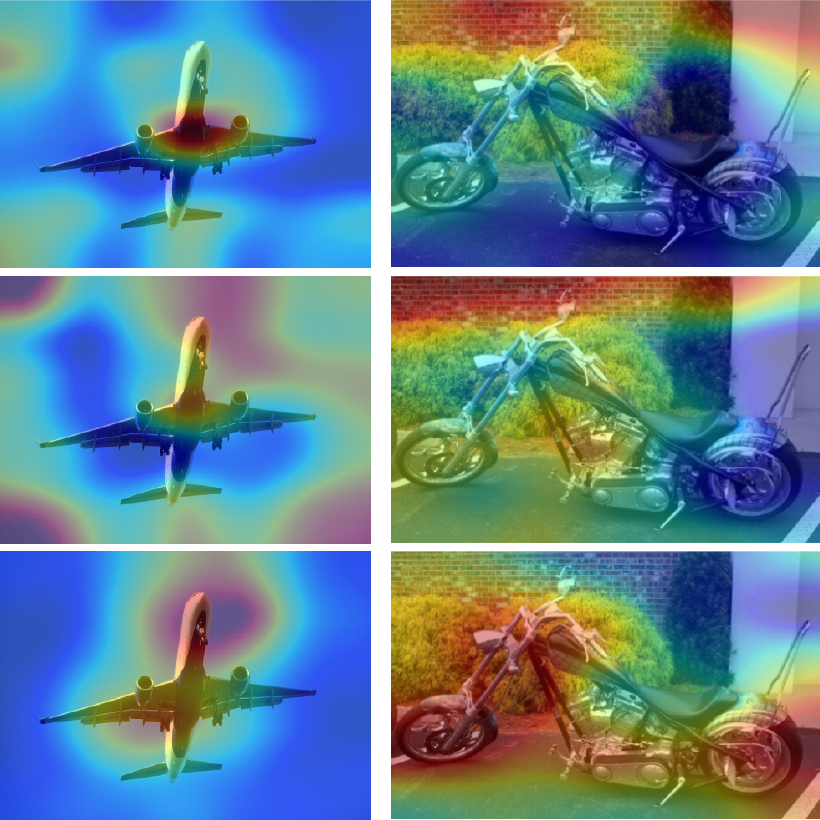}
        \caption{
          Visualized attention of feature pooling methods.
          From top to bottom, rows correspond to NP, GAP, SAP. The red area corresponds to high score for class.
          }
        \label{fig:pooling_gradcam}
    \end{minipage} 
\end{figure}

\subsubsection{Ablation on pooling operators for guidance loss.}
As explained in section~\ref{sec:3.2}, the choice of pooling operator for guidance loss affects the learning. 
We conduct an ablation study on the effect of the pooling operator on VisDA-17 syn-to-real generalization. 
We compare 5 different pooling strategies: no pooling~(NP), global average pooling~(GAP), channel pooling~(CP), spatial pooling~(SP), and self-attention pooling~(SAP), where spatial pooling takes both height pooling and width pooling.
In Fig.~\ref{fig:pooling_chart}, we show the generalization performance in VisDA-17 for various pooling strategies.
We also conduct the ablation study on removing causal invariance loss to
identify the effect of the pooling operator on the generalization. 
In both situations, we observe that self-attention pooling outperforms other pooling strategies. 

We also investigate the attention map using Grad-CAM~\cite{selvaraju2017grad} to qualitatively analyze the effect of pooling operators for guidance loss. In Fig.~\ref{fig:pooling_gradcam}, we observe a tendency that SAP comprehensively includes more semantically meaningful regions than NP which only focuses on detailed features with relatively narrow receptive fields. 
Also, one can observe that SAP focuses less on meaningless parts compared to GAP, which does not sufficiently encode informative features with a relatively wide receptive field.

\begin{table}[t]
    \begin{minipage}[t]{.31\linewidth}
        \centering
        \caption{Ablation on the choice of feature stages $S_{\text{G}}$ and $S_{\text{CI}}$.}
        \resizebox{0.85\textwidth}{!}{
            \begin{tabular}{cccc ccc}
            \toprule
                &$S_{\text{G}}$ && $S_{\text{CI}}$ && Acc & \\
            \midrule
                &\{0,1,2\} && \{0,1,2\} && 59.8 &\\
                &\{0,1,2\} && \{3,4\} && 55.9 &\\
                &\{3,4\} && \{0,1,2\} && 66.2 &\\
                &\{3,4\} && \{3,4\} && \textbf{67.5} & \\
            \bottomrule
            \label{table:feature_stages}
            \end{tabular}
        }
    \end{minipage}
    \hfill
    \begin{minipage}[t]{.34\linewidth}
        \centering
        \caption{Ablation on the choice of temperatures $\tau,\bar{\tau}$ for $\mathcal{L}_{\text{CI}}$. }
        \resizebox{0.97\textwidth}{!}{
                \begin{tabular}{cccccc}
                \toprule
                    $\bar{\tau}$ && 0.02 &  0.04 & 0.06  & 0.08 \\
                \midrule
                    $\tau=0.12$ && 66.6 & \textbf{67.5} & 66.5 & 66.3 \\
                \midrule
                \midrule
                    $\tau$ && 0.08 &  0.1 & 0.12  & 0.14 \\
                \midrule
                    $\bar{\tau}=0.04$  && 66.8 & 67.1 & \textbf{67.5} &67.4 \\
                \bottomrule

                \label{table:temperature_scale}
                \end{tabular}
        }
    \end{minipage}%
    \hfill
    \begin{minipage}[t]{.29\linewidth}
        \centering
        \caption{Ablation on the loss functions for style-invariant learning.}
        \resizebox{0.78\textwidth}{!}{
            \begin{tabular}{c l c cc c}
            \toprule
                & Loss & $\mathcal{L}_{\text{G}}$ && Acc & \\
            \midrule
                & InfoNCE  & \xmark && 56.2 & \\
                &          & \cmark && 64.7 & \\
                \midrule
                & \textbf{CI loss} & \xmark && 58.6 & \\
                &                    & \cmark && \textbf{67.5} & \\
            \bottomrule
            \label{table:ablation_contrastive_loss}
            \end{tabular}
        }
    \end{minipage} 
    \hfill
\end{table}



            
\subsubsection{Ablation on feature stages.}
We conduct an ablation study on generalization performance with different feature stage for $\mathcal{L}_{\text{CI}}$ and $\mathcal{L}_{\text{G}}$. 
Let us denote $S_{\text{G}}$ and $S_{\text{CI}}$ to be the set of layers that we used for computation of guidance loss and causal invariance loss.
In ResNet-101, there is one convolutional layer and 4 ResBlocks. 
We grouped the first three layers and last two layers and experimented on 4 combinations of guidance loss and causal invariance loss with different layer groups. 
As shown in Table~\ref{table:feature_stages}, the computation of guidance loss and causal invariance loss with deeper feature stages is more effective in generalization than the shallower one, as the deeper stage of features holds richer semantic information.
This result is consistent with the tendency of deep layers to encode content discriminative information as Pan et al. pointed out in IBN-Net~\cite{Pan_2018_ECCV}. 

\subsubsection{Ablation on temperature scale.}
We conduct an ablation study on the effect of different temperature scales for causal invariance loss. 
In Table~\ref{table:temperature_scale}, when $\tau=0.12$ we observe that small value of $\bar{\tau}$ is beneficial for the performance. 
Conversely, when $\bar{\tau}$ is fixed to 0.04, relatively high value of $\tau$ has better performance.
Thus, we choose $\tau=0.12$ and $\bar{\tau}=0.04$ for our experiments. 

\subsubsection{Ablation on causal invariance loss.}
We conduct an ablation study on the effect of different contrastive losses for causal invariance loss. 
We compare proposed $\mathcal{L}_{\text{CI}}$ and InfoNCE loss~\cite{mocov2} on VisDA-17 image classification.
To see the effect of causal invariance loss solely, we further removed the guidance loss. 
In Table~\ref{table:ablation_contrastive_loss}, we observe that  $\mathcal{L}_{\text{CI}}$ is more effective in syn-to-real-generalization that InfoNCE loss regardless of the  guidance loss.

\setlength{\tabcolsep}{4pt}
\begin{table}[t]
\begin{center}
\caption{Comparison of mIoU~(\%) with existing synthetic-to-real generalization and domain generalization methods on GTAV-Cityscapes datasets. All methods used ResNet-50 as base architecture. }
\label{table:segmentation_miou_gain}
 \resizebox{.98\columnwidth}{!}{
    \begin{tabular}{c|ccccc|c}
    \toprule
     Method & IBN-Net~\cite{Pan_2018_ECCV}  & DRPC~\cite{yue2019domain}  & RobustNet~\cite{Choi_2021_CVPR} & ASG~\cite{asg} 
      & CSG~\cite{csg} & \textbf{GCISG} \\
    \midrule
    No adapt & 22.17  & 32.45 & 28.95  & 23.29 & 25.43 & 26.67 \\
    Adapt & 29.64  & 37.42 & 36.58  & 29.65 & 35.27 & \textbf{39.01} \\
    \midrule
    mIoU $\uparrow$(\%) & 7.47  & 4.97 & 7.63  & 6.36 & 9.84 & \textbf{12.34} \\
    \bottomrule
    \end{tabular}}
\end{center}
\end{table}
\setlength{\tabcolsep}{1.4pt}

\definecolor{cityvoid}{RGB}{0,0,0}
\definecolor{cityroad}{RGB}{128,64,128}
\definecolor{citysidewalk}{RGB}{244,35,232}
\definecolor{citybuilding}{RGB}{70,70,70}
\definecolor{citywall}{RGB}{102,102,156}
\definecolor{cityfence}{RGB}{190,153,153}
\definecolor{citypole}{RGB}{153,153,153}
\definecolor{citytl}{RGB}{250,170,30}
\definecolor{cityts}{RGB}{220,220,0}
\definecolor{cityveg}{RGB}{107,142,35}
\definecolor{cityterrain}{RGB}{152,251,152}
\definecolor{citysky}{RGB}{70,130,180}
\definecolor{cityperson}{RGB}{220,20,60}
\definecolor{cityrider}{RGB}{255,0,0}
\definecolor{citycar}{RGB}{0,0,142}
\definecolor{citytruck}{RGB}{0,0,70}
\definecolor{citybus}{RGB}{0,60,100}
\definecolor{citytrain}{RGB}{0,80,100}
\definecolor{citymotor}{RGB}{0,0,230}
\definecolor{citybicycle}{RGB}{119,11,32}


\begin{figure}[t]
\centering
\includegraphics[width=0.95\linewidth]{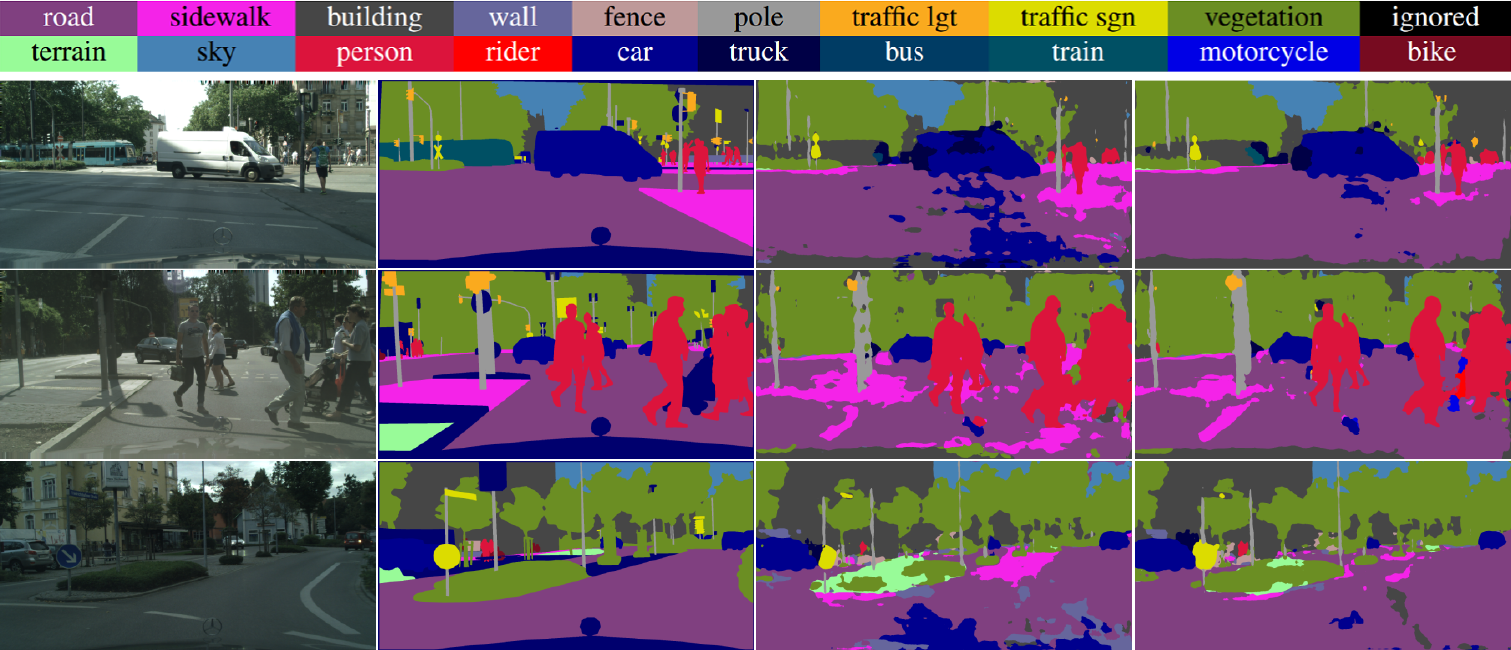}
\caption{
Segmentation results on GTAV to Cityscapes. From left to right,
columns correspond to original images, ground truth, prediction results of CSG, and GCISG.
}
\label{fig:seg_results}
\end{figure}
\subsection{Semantic Segmentation}

\subsubsection{Datasets.}
For semantic segmentation, the source domain is GTAV~\cite{Richter_2016_ECCV} and the target domain is Cityscapes~\cite{Cordts2016Cityscapes}.
GTAV is a large-scale dataset containing 24,996 synthetic driving-scene images with resolution 1914${\times}$1052, taken from the Grand Theft Auto V game engine.
Cityscapes street scene dataset contains 2,975 train images and 500 validation images with resolution 2048${\times}$1024, taken from European cities.
All the images in the source and target dataset have pixel-level annotations with 19 semantic categories.
For evaluation, we report the mean intersection over union~(mIoU) on the Cityscapes validation set.

\subsubsection{Implementation details.}

We adopt DeepLab-V3~\cite{DBLP:journals/corr/ChenPSA17} with ImageNet pre-trained ResNet-50 backbone for our semantic segmentation network. 
We employ an SGD optimizer with a learning rate of 0.001 and a batch size of 8 for 40 epochs.
We crop an image with the size of 512$\times$512 and apply color jittering and multi-scale resizing.

\subsubsection{Results.}
In Table \ref{table:segmentation_miou_gain}, we report the mIoU~(\%) of the semantic segmentation models trained with various syn-to-real generalization methods. 
We observe that the model trained with GCISG achieves the best 
performance gain. 
It is worth noting that IBN-Net~\cite{Pan_2018_ECCV} and RobustNet~\cite{Choi_2021_CVPR}
modifies the normalization blocks,
and DRPC~\cite{yue2019domain} requires preparation steps of stylizing images before training.
In contrast, our method can be applied to any general architecture with no additional preparation steps.
In Fig.~\ref{fig:seg_results}, we qualitatively compare with CSG~\cite{csg} by visualizing the results of segmentation on validation images.

\begin{table}[t]
    \begin{minipage}[t]{.48\linewidth}
        \centering
        \caption{The average precision~(AP) of Sim10k to Cityscapes object detection with various methods.
        All experiments are run by us.}
        \label{table:detection_ap}
        \begin{tabular}{lcccc}
        \toprule
         Method && $\text{AP}$ & ${\text{AP}_{50}}$ & ${\text{AP}_{75}}$ \\
        \midrule
        Baseline             && 24.8  & 45.1 & 24.0 \\
        CSG         && 28.5  & 50.4 & 28.1 \\
        \midrule
        \textbf{GCISG} && \textbf{30.6}  & \textbf{54.6} & \textbf{29.5}  \\
        \bottomrule
        \end{tabular}
    \end{minipage}
    \hfill
    \begin{minipage}[t]{.48\linewidth}
        \centering
        \caption{Top-1 accuracy~(\%) of unsupervised domain adaptation on VisDA-17 by using CBST initialized with different methods.}
        \label{table:uda}
        \begin{tabular}{lcc}
        \toprule
        Method && $\text{Acc}_{\tt{VisDA}}$\\
        \midrule
        Source only + CBST~\cite{Zou_2019_ICCV} && 76.4 \\
        ASG + CBST~\cite{asg}                   && 82.5 \\
        \midrule
        \textbf{GCISG + CBST}            && \textbf{83.6} \\
        \bottomrule
        \end{tabular}
    \end{minipage} 
\end{table}

\begin{figure}[t]
\centering
\includegraphics[width=0.95\linewidth]{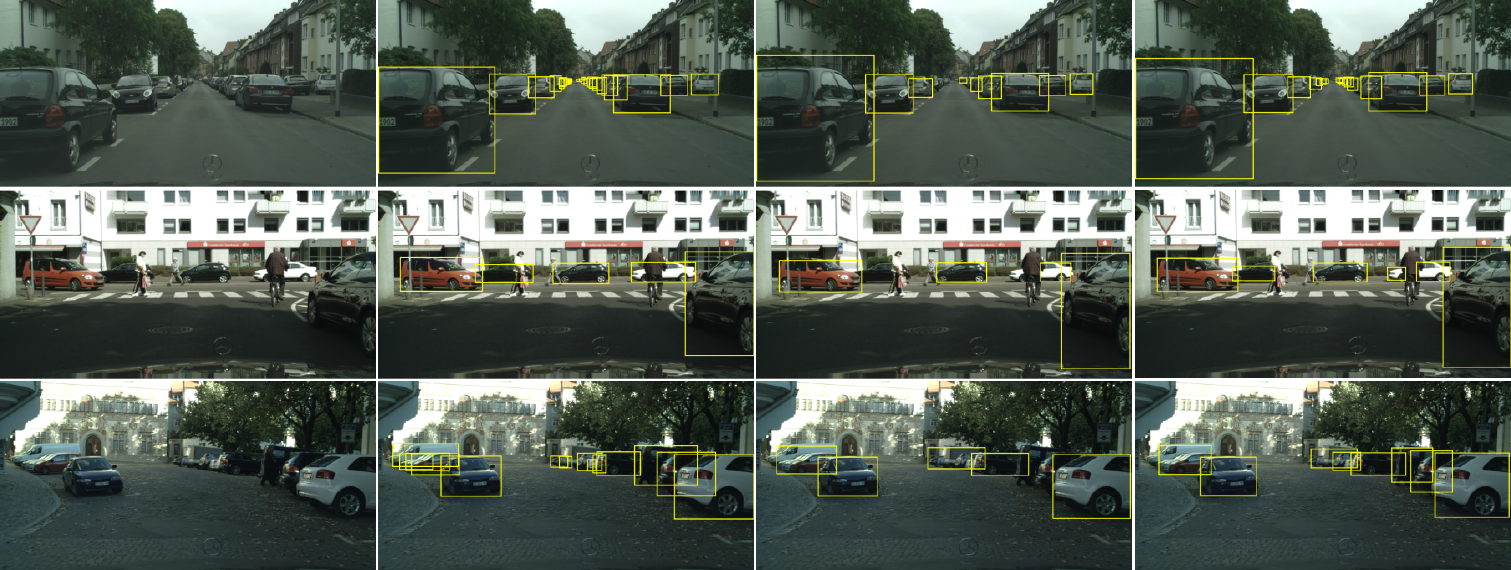}
\caption{
Object detection results on Sim10k to Cityscapes. From left to right,
columns correspond to original images, ground truth, prediction results of CSG, and GCISG.
}
\label{fig:det_results}
\end{figure}
\subsection{Object Detection}

\subsubsection{Datasets.}

We conduct object detection experiments on the source domain of Sim10k~\cite{johnson2017driving} to the target domain of Cityscapes.
Sim10k dataset consists of 10,000 images with bounding box annotations on the car object.
Images in Sim10k are rendered by the Grand Theft Auto V game engine, where the resolution of the images are 1914${\times}$1052.
Since Cityscapes does not contain bounding box labels,
we generate bounding box labels from polygon labels,
following the method of~\cite{chen2018domain}.

\subsubsection{Implementation details.}

We experiment on Faster R-CNN~\cite{7485869}
as the base detector and ImageNet pre-trained ResNet-101 with FPN~\cite{DBLP:journals/corr/LinDGHHB16} as the backbone.
We use an SGD optimizer with a learning rate of 0.001, and a batch size of 4 for 30 epochs.
We use the same hyperparameter settings for the original Faster R-CNN except that we add color jittering for data augmentation.
Since CSG~\cite{csg} does not contain experiments on object detection, we implement the CSG framework for our experiments for a fair comparison.

\subsubsection{Results.}
We evaluate the average precision of bounding boxes
on each generalization method following the COCO~\cite{lin2014microsoft} evaluation protocol.
The results are shown in Table~\ref{table:detection_ap}.
Compared to the previous state-of-the-art
syn-to-real generalization method~\cite{csg},
GCISG achieves around 2.1\% points improvement in AP and
4.2\% points improvement in AP50.
In Fig.~\ref{fig:det_results}, we qualitatively compare with the CSG~\cite{csg} by visualizing the results of the object detection on validation images. 


\subsection{Unsupervised Domain Adaptation}

In this section, we demonstrate the effectiveness of GCISG for the unsupervised domain adaptation~(UDA) task.
We conduct experiments on class-balanced self-training~(CBST)~\cite{Zou_2018_ECCV} framework, where we use the model trained with GCISG as a starting point for the adaptation.

We perform UDA experiments on the VisDA-17 image classification dataset. 
We follow the setting of~\cite{Zou_2019_ICCV}, where we set the starting portion of the pseudo label {\it{p}} to 20\%,
and empirically add 5\% to {\it{p}} for each epoch until it reaches 50\%.
Remark that, unlike previous syn-to-real generalization tasks, we freeze the normalization layers when training the source model, where we empirically found it better for the UDA task. 

In Table \ref{table:uda}, we present an accuracy on VisDA-17 validation dataset under various initialization methods for CBST framework. 
Compared to the baseline, our method remarkably boosts the performance by 7\%, achieving 83.6\%.
Also, the model trained from GCISG outperforms that from ASG by 1\%.

\section{Conclusion}
{
In this work, we present GCISG, which enhances the syn-to-real generalization by learning style-invariant representation and retaining the semantic knowledge of the ImageNet pre-trained model simultaneously.
Through extensive experiments on VisDA-17 image classification, GTAV-Cityscapes semantic segmentation, and object detection, we demonstrate the effectiveness of our method.

Our work shows that we can extract useful information over the style changes that are useful for generalization. 
For future works, we aim to design style transformation methods that can better disentangle the style and content which suits our structural causal model.
We believe that those style transformations can lead to better syn-to-real generalization under our method as they seek better causal invariance.
We leave them for future work. 
}

\subsubsection*{Acknowledgements.}
Kyungmin Lee is supported by Institute of Information $\&$ communications Technology Planning \& Evaluation (IITP) grant funded by the Korea government (MSIT) (No. 2019-0-00075, Artificial Intelligence Graduate School Program (KAIST)).


\clearpage
%
%
\bibliographystyle{splncs04}
\bibliography{egbib}

\end{document}